\documentclass[10pt,twocolumn,letterpaper]{article}

\usepackage{cvpr}              

\usepackage{graphicx}
\usepackage{amsmath}
\usepackage{amssymb}
\usepackage{pifont}
\usepackage[table]{xcolor}
\usepackage{multirow}
\usepackage{wrapfig}
\usepackage[pagebackref,breaklinks,colorlinks,citecolor=teal]{hyperref}
\usepackage{cleveref}
\usepackage{marvosym}
\usepackage[most]{tcolorbox}

\usepackage{booktabs}
\usepackage{xcolor}
\usepackage{algorithm}
\usepackage{algpseudocode}
\usepackage{float}

\def\cvprPaperID{XXX} 
\def\confName{CVPR}
\def\confYear{2023}

\begin{document}

\twocolumn[{%
\renewcommand\twocolumn[1][]{#1}%


\title{Fuse4Seg: Image Fusion for Multi-Modal Medical Segmentation \\ via Bi-level Optimization}
\author{
    Yuchen Guo$^{1}$, \quad
    Junli Gong$^{2}$, \quad
    Hongmin Cai$^{3}$, \quad
    Yiu-ming Cheung$^{4}$, \quad \vspace{0.2cm}
    Weifeng Su$^{5}$ \\
    $^{1}$Northwestern University \quad
    $^{2}$Northeastern University \quad
    $^{3}$South China University of Technology \\
    $^{4}$Hong Kong Baptist University \quad
    $^{5}$Beijing Normal - Hong Kong Baptist University \\ \vspace{0.15cm}
    {\tt\small yuchenguo2027@u.northwestern.edu, wfsu@bnbu.edu.cn}
}

\maketitle

\begin{center}
 \centering
 \vspace{-0.28cm}
 \includegraphics[width=1\linewidth]{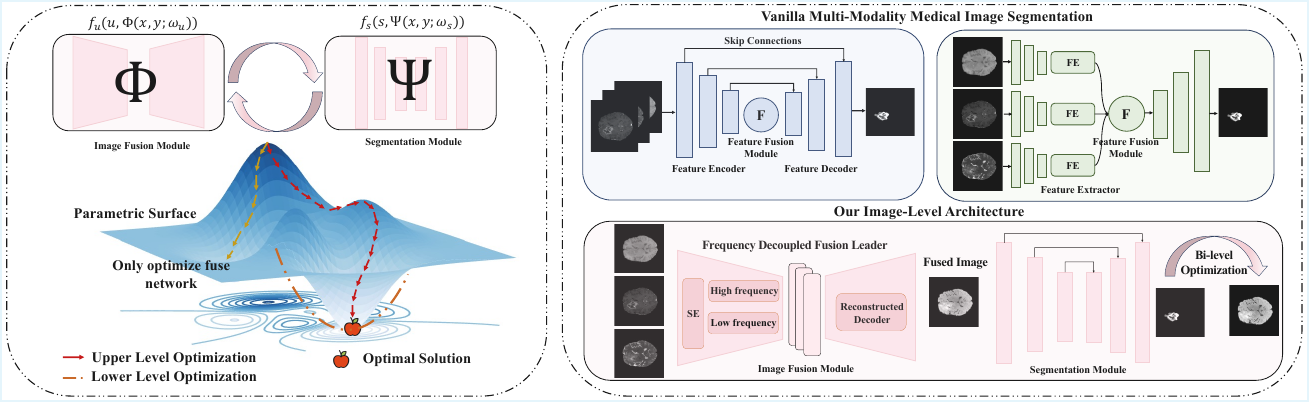}
 \vspace{-6.1mm}
 \captionof{figure}{
 (left) The bi-level optimization learning process with fusion task as the leader and segmentation task as follower. (right) Existing Multi-Modality Medical Image Segmentation Methods \textit{vs.} Our Fuse4Seg.}
 \vspace{.3cm}
 \label{fig:intro}
\end{center}%
}]

\begin{abstract}
Multi-modal medical image fusion is traditionally optimized for human visual perception, aiming to maximize generic contrast and structural fidelity. However, when these visually pleasing fused images are deployed in automated clinical workflows, this visual-semantic discrepancy causes task-agnostic feature degradation, inadvertently smoothing out critical, high-frequency tumor boundaries. To bridge this semantic gap, we propose \textbf{\textit{Fuse4Seg}}, a novel framework that reformulates multi-modal fusion as a cooperative bi-level optimization problem with medical segmentation. Rather than relying on rigid visual metrics, our fusion leader dynamically updates its feature extraction strategy driven directly by semantic gradients backpropagated from the downstream segmentation follower. To guarantee robust physical fidelity alongside semantic utility, we design a frequency-decoupled architecture stringently regularized by a Frequency Decomposition Loss and a Spatial Gradient Loss. This explicit physical anchor prevents anatomical distortion and ensures the lossless preservation of task-critical details. Extensive experiments demonstrate that our task-aware, single-channel fused prior generalizes seamlessly across diverse multi-scale modalities. More impressively, it remarkably surpasses contemporary dual-channel segmentation state-of-the-arts while explicitly providing a readable, "glass-box" physical image to foster clinical visual interpretability and trust.
\end{abstract}

\vspace{-0.3cm}
\section{Introduction}
\label{sec:intro}

Multi-modal Medical Image Fusion (MIF) plays an indispensable role in modern clinical diagnostics\cite{james2014medical, azam2022review}. By integrating the complementary physical properties of diverse imaging modalities—such as combining the high spatial resolution of anatomical T1ce MRI with the distinct lesion contrast of FLAIR, or merging MRI with functional PET/SPECT—fusion models generate comprehensive representations that significantly aid human interpretation \cite{tan2025multimodal}. Recently, advanced generative architectures, including Diffusion models and state space models (e.g., Mamba), have pushed MIF to unprecedented heights, generating fused images with remarkable visual fidelity and structural similarity\cite{zhao2023ddfm, xie2024fusionmamba}.

Despite their visual appeal, modern clinical workflows increasingly rely on automated downstream analysis, such as tumor segmentation, to provide objective, quantitative assessments. Herein lies a critical bottleneck: existing MIF techniques are predominantly ported from natural image domains and are rigorously optimized for \textit{human visual perception} (maximizing generic contrast or global entropy)\cite{tang2022matr,li2025bsafusion}. They operate completely disjointly from downstream machine vision tasks. Consequently, when these visually pleasing fused images are fed into segmentation networks, critical task-specific high-frequency gradients—such as subtle, highly irregular tumor boundaries—are often inadvertently smoothed out or overshadowed by synthetic textures. This phenomenon is task-agnostic feature degradation, which stems from a fundamental \textbf{visual-semantic discrepancy}: \textit{the objective misalignment between low-level visual reconstruction and high-level scene understanding.}

To fundamentally bridge this semantic gap, we present \textit{Fuse4Seg}. Instead of treating fusion and segmentation as isolated pipelines, we formulate their intertwined dependency as a cooperative bi-level optimization problem. In this hierarchy, the image fusion network acts as the upper-level leader, while the segmentation network acts as the lower-level follower. Rather than blindly maximizing generic visual metrics, the fusion leader dynamically updates its parameters driven directly by the semantic gradients backpropagated from the downstream follower. This forces the fusion network to intelligently compress multi-modal information into a single-channel, \textit{task-aware prior} that explicitly maximizes downstream segmentation utility while discarding redundant physical noise.

Our fusion leader employs a frequency-decoupled architecture, utilizing Transformers \cite{vaswani2017attention} to model macroscopic tissue contrast and Invertible Neural Networks (INNs) \cite{ardizzone2018analyzing} to losslessly preserve sharp pathological margins. These extractors are stringently regularized by a Frequency Decomposition Loss and a Spatial Gradient Loss to ensure the task-driven fusion strictly preserves anatomical structures rather than collapsing into adversarial noise. Crucially, in stark contrast to contemporary multi-channel segmentation pipelines that process raw modalities through uninterpretable "black-box" latent spaces, our framework explicitly bottlenecks the multi-modal information into a readable, single-channel "glass-box" physical image. This paradigm shift not only alleviates the computational burden on the downstream network but also provides crucial visual interpretability, allowing clinicians to transparently verify the biological basis of the automated diagnosis. Based on the above, our main contributions are summarized:
\begin{itemize}
    \item We propose \textit{Fuse4Seg}, a novel framework that reformulates multi-modal medical image fusion as a cooperative bi-level optimization problem, advancing the paradigm from human-perception-driven visual enhancement to task-driven semantic fusion.
    \item We design a physically-anchored joint architecture. By integrating a Frequency Decomposition Loss and a Spatial Gradient constraint, we prevent adversarial degradation under semantic pressure, perfectly balancing anatomical fidelity (physics) with diagnostic accuracy (semantics).
    \item Extensive experiments demonstrate that our single-channel prior remarkably surpasses contemporary dual-channel segmentation SOTAs across diverse modalities. Crucially, bottlenecking abstract features into a physical image provides ``glass-box'' visual interpretability, fostering essential clinical trust.
\end{itemize}

\section{Related Works}
\subsection{Medical Image Segmentation}
Medical image segmentation is pivotal for delineating anatomical and pathological structures\cite{azad2024medical, muller2022towards, wang2022medical}. While early advancements were driven by CNN-based U-shaped architectures \cite{huang2020unet, chen2017deeplab} and later revolutionized by Transformer-based models for global context modeling (e.g., TransUNet \cite{chen2021transunet}, Swin-Unet \cite{cao2022swin}, and Mamba-based variants \cite{wang2024mamba}), modern clinical scenarios increasingly rely on \textit{multi-modal} inputs. This dominant paradigm simply employs channel-wise stacking at the input level (e.g., nnU-Net \cite{isensee2021nnu}). However, this brute-force feature-level fusion heavily inflates computational overhead and operates as a "black-box," entirely neglecting the interpretable, holistic physical priors that a unified fused image could provide.

\subsection{Medical Image Fusion}
Existing image fusion methods primarily include autoencoder-based models \cite{zhao2023cddfuse, guo2025dae, zhao2020didfuse}, GANs \cite{ma2020ddcgan, liu2022target}, diffusion models \cite{zhao2023ddfm, tang2025mask, yi2024diff}, and recent Mamba architectures \cite{xie2024fusionmamba}. Most of these techniques are rigidly optimized for \textit{human visual perception} (maximizing generic structural similarity)\cite{zhao2024equivariant, liu2021learning}. When deployed in automated diagnostic pipelines, this visual-centric objective inherently causes task-agnostic feature degradation. While some recent natural image fusion models have begun incorporating downstream losses heuristically \cite{liu2023multi, liu2022target}, medical domains lack a rigorous mathematical framework to align fusion with complex semantic targets. Our \textit{Fuse4Seg} successfully bridges this critical gap in MIF by formulating the interaction as a cooperative bi-level optimization problem, explicitly generating a task-driven physical prior.

\section{Methodology}
\subsection{Bi-level Optimization Formulation}
Medical Image Fusion (MIF) is conventionally treated as an isolated preprocessing step prior to Medical Image Segmentation (MIS). However, optimizing fusion independently often leads to the loss of critical semantic features required by downstream tasks. To bridge this gap, we model the joint fusion and segmentation process as a Stackelberg game, where the fusion module acts as the leader (upper task) and the segmentation module acts as the follower (lower task). 

Let $x_1, x_2 \in \mathbb{R}^{H \times W}$ denote the input medical images from two different modalities (e.g., T1ce and FLAIR). We define the fusion network as $\Phi$ parameterized by $\theta_f$, which generates the single-channel fused image $x_{f} = \Phi(x_1, x_2; \theta_f)$. Subsequently, the segmentation network $\Psi$, parameterized by $\theta_s$, takes $x_{f}$ as input to predict the semantic mask $\hat{y} = \Psi(x_{f}; \theta_s)$.

The interdependence between fusion and segmentation can be mathematically formulated as a bi-level optimization problem:
\begin{equation}
\begin{aligned}
    \min_{\theta_f} \quad & \mathcal{L}_{upper}(\theta_f, \theta_s^*) \\
    & = \mathcal{L}_{seg}\Big(\Psi(\Phi(x_1, x_2; \theta_f); \theta_s^*), y\Big) + \lambda \mathcal{L}_{fuse}(\theta_f), \\
    \text{s.t.} \quad & \theta_s^*(\theta_f) = \arg\min_{\theta_s} \mathcal{L}_{lower}(\theta_s, \theta_f) \\
    & = \mathcal{L}_{seg}\Big(\Psi(\Phi(x_1, x_2; \theta_f); \theta_s), y\Big).
\end{aligned}
\label{eq:bilevel}
\end{equation}
where $\mathcal{L}_{seg}$ represents the task-driven segmentation loss, and $\mathcal{L}_{fuse}$ is an abstract physical regularity term preserving modality-specific textures and anatomical boundaries. To eliminate ambiguity, in practical implementation, this regularization is instantiated as a combination of decomposition and spatial gradient constraints, defined as $\lambda \mathcal{L}_{fuse}(\theta_f) \triangleq \alpha \mathcal{L}_{decomp} + \beta \mathcal{L}_{grad}$. Unlike previous disjointed methods, Eq. \ref{eq:bilevel} enforces that the optimal fusion parameters $\theta_f$ are explicitly driven by the performance of the optimal segmentation network $\theta_s^*$.

where $\mathcal{L}_{seg}$ represents the task-driven segmentation loss, and $\mathcal{L}_{fuse}$ is an abstract physical regularity term preserving modality-specific textures and anatomical boundaries. To eliminate ambiguity, in practical implementation, this regularization is instantiated as a combination of decomposition and spatial gradient constraints, defined as $\lambda \mathcal{L}_{fuse}(\theta_f) \triangleq \alpha \mathcal{L}_{decomp} + \beta \mathcal{L}_{grad}$. Unlike previous disjointed methods, Eq. \ref{eq:bilevel} enforces that the optimal fusion parameters $\theta_f$ are explicitly driven by the performance of the optimal segmentation network $\theta_s^*$.

\subsection{First-Order Cooperative Training Strategy}
Solving the exact bi-level optimization in Eq. \ref{eq:bilevel} is computationally intractable for deep neural networks due to the requirement of calculating the optimal $\theta_s^*$ for every fusion update. To efficiently solve this while preventing the framework from collapsing into adversarial noise, we design an advanced, empirically stable first-order alternating strategy. This strategy features a two-stage training paradigm: a Follower Warm-up stage and an Asymmetric Bi-level Rollout stage, regularized by a strict patient-level data-splitting protocol.

\textbf{Stage 1: Follower Warm-up.}
Initializing both the fusion leader and segmentation follower from scratch simultaneously inevitably leads to optimization divergence. To establish a stable semantic baseline, we freeze the fusion network and utilize a deterministic physical prior (e.g., the average of the multi-modal inputs, $x_{avg} = \frac{1}{2}(x_1 + x_2)$) to train the segmentation follower for an initial $E_{warm}$ epochs. This guarantees the follower possesses rudimentary tumor-localization capabilities before providing semantic guidance.

\begin{algorithm}[t]
\caption{Cooperative Training Strategy}
\label{alg:bilevel}
\textbf{Input:} Init. fusion params $\theta_f$, seg. params $\theta_s$, EMA $\hat{\theta}_f \leftarrow \theta_f$. \\
\hspace*{3.2em} Multi-modal dataset $\mathcal{D}$, inner steps $K$, decay $m$. \\
\textbf{Output:} Optimized EMA fusion $\hat{\theta}_f$, seg. params $\theta_s^*$.
\begin{algorithmic}[1]
\Statex \textcolor{green!50!black}{\# \textit{Stage 1: Follower Warm-up}}
\For{$epoch = 1$ \textbf{to} $E_{warm}$}
    \State Freeze $\theta_f$. Generate prior $x_{avg} = \frac{1}{2}(x_1 + x_2)$.
    \State Update $\theta_s$ using $\mathcal{L}_{seg}\big(\Psi(x_{avg}; \theta_s), y\big)$.
\EndFor

\Statex \textcolor{green!50!black}{\# \textit{Stage 2: Asymmetric Bi-level Rollout}}
\While{not converged}
    \State Sample batch $\mathcal{B}$, split into $\mathcal{D}_{tr}, \mathcal{D}_{val}$ (patient-level).
    
    \Statex \indent \textcolor{green!50!black}{\# \textit{Inner Loop: Follower Update}}
    \State Unfreeze $\theta_s$. Freeze $\theta_f$.
    \For{$k = 1$ \textbf{to} $K$}
        \State Fused images on $\mathcal{D}_{tr}$: $x_f^{(tr)} = \Phi(x_1, x_2; \theta_f)$.
        \State $\theta_s \leftarrow \theta_s - \eta_s \nabla_{\theta_s} \mathcal{L}_{seg}\big(\Psi(x_f^{(tr)}; \theta_s), y^{(tr)}\big)$.
    \EndFor
    
    \Statex \indent \textcolor{green!50!black}{\# \textit{Outer Loop: Leader Update (Semantic Guided)}}
    \State Freeze $\theta_s$. Unfreeze $\theta_f$.
    \State Fused images on $\mathcal{D}_{val}$: $x_f^{(val)} = \Phi(x_1, x_2; \theta_f)$.
    \State Compute upper loss with physical anchor (Eq. 4):
    \State \quad $\mathcal{L}_{upper} = \mathcal{L}_{seg}\big|_{\mathcal{D}_{val}} \!+ \alpha \mathcal{L}_{decomp} \!+ \beta \mathcal{L}_{grad} \!+ \gamma \mathcal{L}_{recon}$.
    \State Update fusion net: $\theta_f \leftarrow \theta_f - \eta_f \nabla_{\theta_f} \mathcal{L}_{upper}$.
    \State Update EMA params: $\hat{\theta}_f \leftarrow m \hat{\theta}_f + (1 - m) \theta_f$.
\EndWhile
\end{algorithmic}
\end{algorithm}

\begin{figure*}[htbp]
\centering
\includegraphics[width=\textwidth]{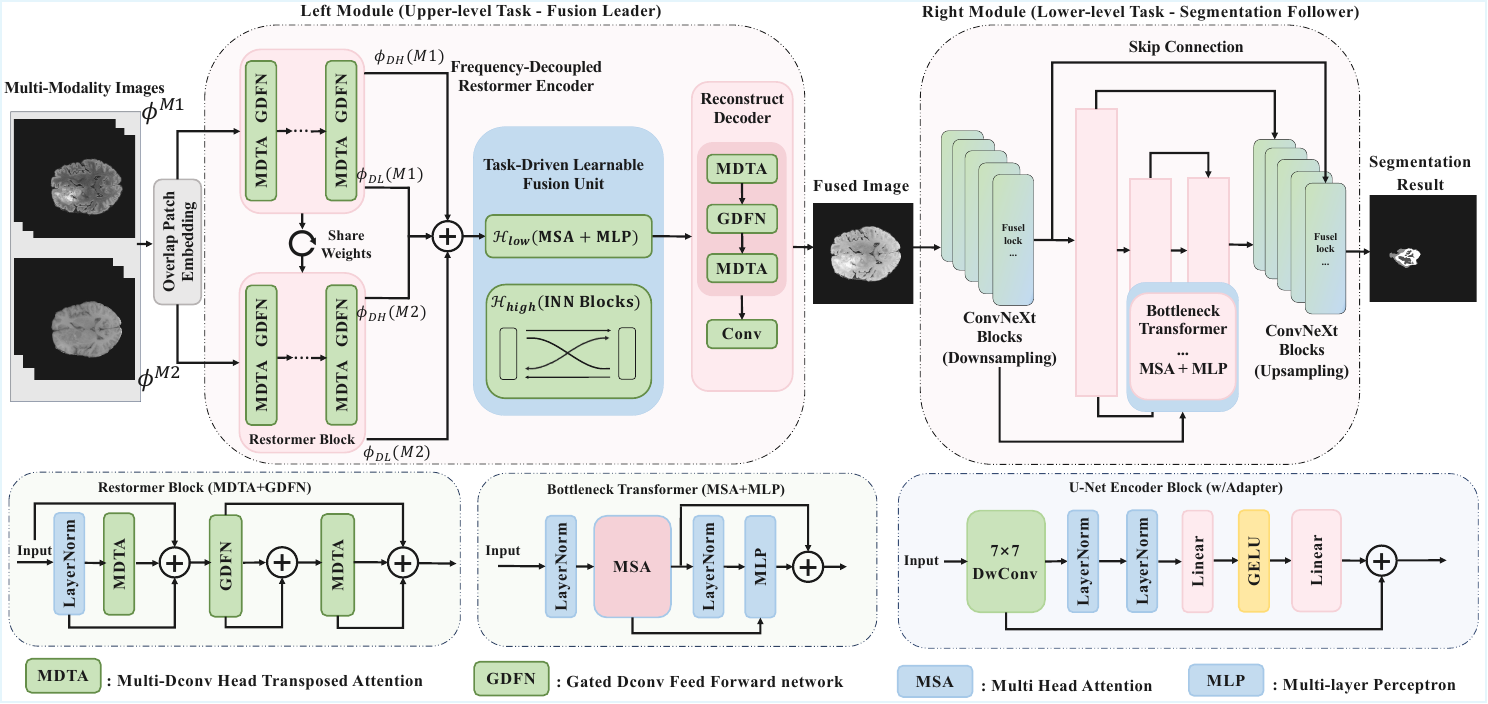} 
\caption{The overall framework of our Fuse4Seg, which consist of a fusion module and a segmentation module.}
\label{fig:model}
\end{figure*}

\textbf{Stage 2: Asymmetric Bi-level Rollout.}
After warm-up, we activate the bi-level cooperative loop. Within each batch $\mathcal{B}$, we partition the data into two disjoint subsets at the patient-level: $\mathcal{D}_{tr}$ for training the follower and $\mathcal{D}_{val}$ for updating the leader. 

\textit{Inner Loop (Follower Update):} Theoretical bi-level optimization assumes the follower achieves a local optimum $\theta_s^*$ given the current leader $\theta_f$. To closely approximate this, we employ an \textbf{asymmetric update scheme}. For every fusion update, the segmentation network performs $K$ consecutive gradient descents on $\mathcal{D}_{tr}$:
\begin{equation}
\begin{aligned}
    \theta_s^{(k)} \leftarrow {} & \theta_s^{(k-1)} \\
    & - \eta_s \nabla_{\theta_s} \mathcal{L}_{seg}\Big(\Psi(\Phi(x_1, x_2; \theta_f); \theta_s^{(k-1)}), y\Big) \Big|_{\mathcal{D}_{tr}}, \\
    & k \in \{1, \dots, K\},
\end{aligned}
\end{equation}

\textit{Outer Loop (Leader Update):} After $K$ inner steps, we freeze $\theta_s$ and update the fusion network on $\mathcal{D}_{val}$. Driven by the semantic gradients backpropagated through the optimized follower, the leader dynamically refines its fusion strategy. To prevent the fusion image from collapsing into task-overfitted adversarial artifacts, we explicitly enforce a lightweight physical anchor via a reconstruction loss ($\mathcal{L}_{recon}$). The upper-level objective simplifies to:
\begin{equation}
    \mathcal{L}_{upper} = \mathcal{L}_{seg}\Big|_{\mathcal{D}_{val}} + \alpha \mathcal{L}_{decomp} + \beta \mathcal{L}_{grad} + \gamma \mathcal{L}_{recon}.
\end{equation}
The parameter $\theta_f$ is then updated via direct partial derivatives. Furthermore, to enhance the physical stability of the generated fusion prior during clinical inference, we maintain an Exponential Moving Average (EMA) of the fusion network weights:
\begin{equation}
    \hat{\theta}_f \leftarrow m \hat{\theta}_f + (1 - m) \theta_f,
\end{equation}
where $m$ is the momentum decay rate. The complete training procedure is detailed in Algorithm \ref{alg:bilevel}.

\subsection{Fusion Module}
\subsubsection{Frequency-Decoupled Encoder} 
In multi-modal neuroimaging, diagnostic information is intrinsically stratified across spatial frequencies. Low-frequency components typically govern macroscopic anatomical topology and global tissue contrast, and high-frequency signals encapsulate crucial modality-specific pathological signatures, such as sharp enhancing tumor margins, diffuse edema boundaries, and intricate micro-vascular textures. To explicitly leverage this biological property and prevent the feature entanglement of healthy structural background with subtle lesion details, our encoder first processes the source images through shared Restormer blocks to establish a robust global representation. Subsequently, to avoid task-agnostic feature degradation, the network branches into two specialized frequency-oriented extractors.

For low-frequency features ($F_{low}^1, F_{low}^2$) representing macroscopic structures and base intensity, we employ standard Multi-Head Self-Attention (MSA) and MLP layers. Conversely, for high-frequency features ($F_{high}^1, F_{high}^2$) capturing critical diagnostic edges (e.g., subtle tumor boundaries), we strictly prevent information loss by employing Invertible Neural Network (INN) blocks. Following the architectural design of CDDFuse \cite{zhao2023cddfuse}, the INN block utilizes Haar wavelet transformations to squeeze spatial dimensions, followed by affine coupling layers. Specifically, the intermediate feature $z$ is split along the channel dimension into $z_1$ and $z_2$, undergoing sequential transformations:
\begin{equation}
    z_1' = z_1 \odot \exp(\theta_{\rho}(z_2)) + \theta_{\eta}(z_2), \quad z_2' = z_2 + \theta_{\phi}(z_1'),
\end{equation}
where $\odot$ denotes element-wise multiplication, and $\theta_{\rho}$, $\theta_{\eta}$, and $\theta_{\phi}$ represent arbitrary non-linear functions modeled by dense convolutional blocks \cite{zhao2023cddfuse}. The output is then concatenated as $z' = [z_1', z_2']$. This rigorously invertible design guarantees zero-loss preservation of high-frequency pathological gradients during the forward pass.

\subsubsection{Learnable Fusion Unit} 
To seamlessly integrate the decoupled features under the guidance of downstream semantic gradients, we discard static arithmetic rules in favor of fully learnable extraction modules ($\mathcal{H}_{low}$ and $\mathcal{H}_{high}$). The frequency-specific features are first aggregated via element-wise addition and then dynamically optimized:
\begin{equation}
\begin{aligned}
    F_{low}^{fused} &= \mathcal{H}_{low}(F_{low}^1 + F_{low}^2), \\
    F_{high}^{fused} &= \mathcal{H}_{high}(F_{high}^1 + F_{high}^2).
\end{aligned}
\end{equation}
Here, $\mathcal{H}_{low}$ employs MSA and MLP layers to establish optimal global contrast, while $\mathcal{H}_{high}$ utilizes sequential INN blocks to implicitly learn the optimal spatial weighting for high-frequency pathological details. This data-driven aggregation ensures that the feature selection process adapts strictly to the semantic pressure from the segmentation follower. Finally, an anchor-free Transformer decoder reconstructs the single-channel fused image $x_f$ from the concatenated features $\{F_{low}^{fused}, F_{high}^{fused}\}$. To guarantee physical validity and stabilize downstream segmentation, $x_f$ is strictly constrained to $[0, 1]$ domain via Min-Max normalization.

\subsection{Segmentation Module}
To efficiently decode the task-driven fused prior $x_f$, our segmentation module employs a Hybrid CNN-Transformer U-shape architecture. The hierarchical backbone utilizes ConvNeXt blocks \cite{woo2023convnext} with large-kernel ($7 \times 7$) depthwise convolutions, naturally capturing expansive heterogeneous pathologies (e.g., peritumoral edema) with minimal parameter overhead. To preserve fine-grained spatial details, symmetrical skip connections fuse high-resolution features directly into the decoder. Crucially, a Transformer Bottleneck with Multi-Head Self-Attention (MSA) is embedded at the lowest resolution to model long-range global semantic dependencies. This hybrid design optimally balances precise anatomical boundary localization with comprehensive context understanding, ultimately providing the robust semantic gradients necessary to guide our upstream fusion leader.

\subsection{Objective Functions}
\subsubsection{Lower-Level Objective (Segmentation Follower)}
In the inner loop of the bi-level optimization, the segmentation network (parameterized by $\theta_s$) acts as the follower. Its sole objective is to achieve optimal tumor delineation given the current fused prior generated by the leader. The lower-level objective $\mathcal{L}_{lower}$ is formulated using a combination of Cross-Entropy (CE) loss and a weighted multi-class Dice loss. To address the severe class imbalance among the background, necrotic core, peritumoral edema, and enhancing tumor, the Dice loss heavily penalizes misclassifications in the foreground:
$$ \mathcal{L}_{Dice} = 1 - \frac{1}{C} \sum_{c=1}^{C} w_c \frac{2 \sum_{i} p_{c,i} y_{c,i} + \epsilon}{\sum_{i} p_{c,i} + \sum_{i} y_{c,i} + \epsilon}, $$
where $p_{c,i}$ and $y_{c,i}$ denote the predicted probability and ground truth for class $c$ at pixel $i$, $\epsilon$ is a smoothing factor, and $w_c$ represents the class-specific penalty weight. The total lower-level objective is defined as:
$$ \mathcal{L}_{lower}(\theta_s; \theta_f) = \mathcal{L}_{Dice} + \mathcal{L}_{CE}. $$
During the asymmetric inner steps, $\theta_s$ is optimized exclusively by minimizing $\mathcal{L}_{lower}$ on the training split $\mathcal{D}_{tr}$.

\subsubsection{Upper-Level Regularizations}
For the outer loop, the fusion network (parameterized by $\theta_f$) acts as the leader. While its primary driving force is the semantic validation loss backpropagated from the optimized follower, updating solely based on this can lead to task-overfitted adversarial artifacts. Therefore, we introduce three complementary physical regularizations:

\textbf{Frequency Decomposition Loss ($\mathcal{L}_{decomp}$):} To explicitly enforce the separation of modality-shared structure and modality-specific details, we use a Correlation Coefficient ($CC$)-based penalty:
$$ \mathcal{L}_{decomp} = \frac{[CC(F_{high}^1, F_{high}^2)]^2}{1.01 + CC(F_{low}^1, F_{low}^2)}. $$
This objective aggressively minimizes the correlation of high-frequency textures while maximizing the correlation of low-frequency anatomy.

\textbf{Spatial Gradient Loss ($\mathcal{L}_{grad}$):} To ensure the sharpest diagnostic boundaries from source modalities are retained, we apply a gradient-based penalty using the Sobel operator $\nabla$:
$$ \mathcal{L}_{grad} = \big\| \nabla x_f - \max(\nabla x_1, \nabla x_2) \big\|_1. $$

\begin{figure*}[t]
    \centering
    \includegraphics[width=\linewidth]{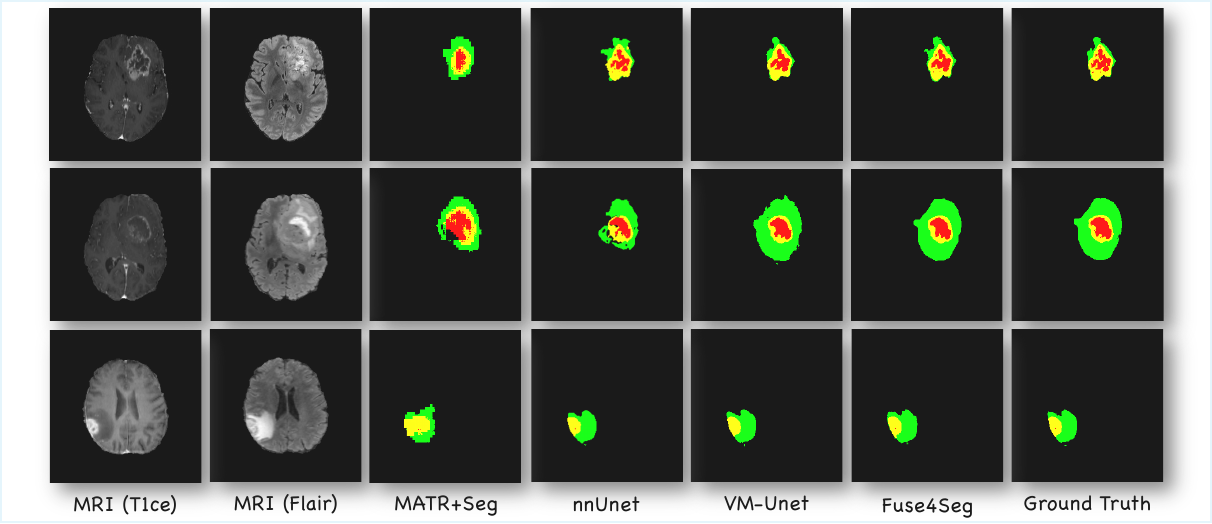} 
    \caption{Qualitative comparison with other SOTA medical segmentation methods.}
    \label{fig:qualitative}
\end{figure*}

\textbf{Physical Reconstruction Anchor ($\mathcal{L}_{recon}$):} To guarantee the clinical readability of the fused image and prevent catastrophic physical collapse during the task-driven updates, we introduce a lightweight Mean Squared Error (MSE) anchor:
$$ \mathcal{L}_{recon} = \big\| x_f - \frac{1}{2}(x_1 + x_2) \big\|_2^2. $$

\subsubsection{Total Upper-Level Objective (Fusion Leader)}
Ultimately, the fusion leader is optimized on the validation split $\mathcal{D}_{val}$ by minimizing the comprehensive upper-level objective. This formulation elegantly balances the downstream semantic guidance (evaluated on the optimal follower $\theta_s^*$) with the aforementioned physical constraints:
\begin{equation*}
\begin{aligned}
    \mathcal{L}_{upper}(\theta_f) = {} & \mathcal{L}_{lower}(\theta_s^*) \Big|_{\mathcal{D}_{val}} + \alpha \mathcal{L}_{decomp} \\
    & + \beta \mathcal{L}_{grad} + \gamma \mathcal{L}_{recon},
\end{aligned}
\end{equation*}
where $\alpha$, $\beta$, and $\gamma$ are empirical hyperparameters strictly controlling the trade-off between task-driven performance and physical image fidelity.

\section{Experiments}
\subsection{Experimental Setup}
Our framework is implemented in PyTorch and trained on a single NVIDIA A100 GPU. We first conduct a Follower Warm-up stage for 40 epochs, where the segmentation network is solely optimized using a deterministic physical prior (i.e., the average of the multi-modal inputs) to establish a stable semantic baseline. Subsequently, the asymmetric bi-level rollout stage is executed for 120 epochs. During this stage, to ensure the follower closely approximates the local optimum, the segmentation network performs $K=5$ inner gradient updates for every single outer update of the fusion leader. Both networks are optimized using the AdamW optimizer with a weight decay of $1 \times 10^{-5}$, coupled with a cosine annealing learning rate scheduler. The initial learning rates for the fusion leader and segmentation follower are empirically set to $1 \times 10^{-4}$ and $3 \times 10^{-4}$, respectively. To maintain a batch size of 8, we leverage Automatic Mixed Precision (AMP) training. For the upper-level objective, the optimization trade-off parameters are set as $\alpha=1.0$, $\beta=5.0$, and $\gamma=0.05$ (for the lightweight physical reconstruction anchor). Additionally, the Exponential Moving Average (EMA) momentum decay rate for stabilizing the fusion leader's weights is set to $m=0.99$.

\begin{table*}[htbp]
\centering
\caption{Quantitative segmentation results on the BraTS 2021 dataset utilizing T1ce and FLAIR modalities. All disjoint fusion baselines are evaluated using our identical segmentation backbone. The best results are highlighted in \textbf{bold}, and the second-best are \underline{underlined}.}
\label{tab:seg_results}
\resizebox{\textwidth}{!}{
\begin{tabular}{l|cc|cc|cc|cc}
\toprule
\multirow{2}{*}{\textbf{Method}} & \multicolumn{2}{c|}{\textbf{NCR} {\scriptsize (Necrotic Core)}} & \multicolumn{2}{c|}{\textbf{ED} {\scriptsize (Peritumoral Edema)}} & \multicolumn{2}{c|}{\textbf{ET} {\scriptsize (Enhancing Tumor)}} & \multicolumn{2}{c}{\textbf{Average}} \\
\cmidrule(lr){2-3} \cmidrule(lr){4-5} \cmidrule(lr){6-7} \cmidrule(lr){8-9}
& \textbf{Dice $\uparrow$} & \textbf{IoU $\uparrow$} & \textbf{Dice $\uparrow$} & \textbf{IoU $\uparrow$} & \textbf{Dice $\uparrow$} & \textbf{IoU $\uparrow$} & \textbf{Mean Dice $\uparrow$} & \textbf{Mean IoU $\uparrow$} \\
\midrule
\multicolumn{9}{c}{\textit{Disjoint Fusion-Segmentation Pipelines}} \\
\midrule
DDFM + Seg & 0.765 & 0.665 & 0.785 & 0.675 & 0.800 & 0.710 & 0.783 & 0.683 \\
BSAFusion + Seg & 0.775 & 0.675 & 0.795 & 0.685 & 0.810 & 0.720 & 0.793 & 0.693 \\
MATR + Seg & 0.780 & 0.685 & 0.805 & 0.690 & 0.820 & 0.730 & 0.801 & 0.701 \\
TFS-Diff + Seg & 0.785 & 0.695 & 0.815 & 0.700 & 0.835 & 0.745 & 0.811 & 0.713 \\
CDDFuse + Seg & 0.795 & 0.700 & 0.825 & 0.705 & 0.840 & 0.750 & 0.820 & 0.718 \\
FusionMamba + Seg& 0.802 & 0.710 & 0.835 & 0.715 & 0.850 & 0.760 & 0.829 & 0.728 \\
\midrule
\multicolumn{9}{c}{\textit{Direct Multi-Channel Segmentation}} \\
\midrule
transU-Net & 0.820 & 0.730 & 0.880 & 0.745 & 0.905 & 0.820 & 0.868 & 0.765 \\
nnU-Net & 0.810 & 0.724 & 0.895 & 0.750 & 0.912 & 0.824 & 0.872 & 0.766 \\
CENet (2025) & 0.835 & 0.740 & 0.899 & 0.759 & 0.912 & 0.832 & 0.882 & 0.777 \\
VM-UNet & \underline{0.875} & \underline{0.798} & \textbf{0.912} & \underline{0.826} & \underline{0.928} & \underline{0.870} & \underline{0.905} & \underline{0.831} \\
\midrule
\multicolumn{9}{c}{\textit{Proposed Method}} \\
\midrule
\textbf{Fuse4Seg (Ours)} & \textbf{0.884} & \textbf{0.807} & \underline{0.908} & \textbf{0.833} & \textbf{0.937} & \textbf{0.884} & \textbf{0.910} & \textbf{0.841} \\
\bottomrule
\end{tabular}
}
\end{table*}

\subsection{Medical Image Segmentation Results}
\paragraph{\textbf{Dataset.}} We evaluate our framework on the BraTS 2021 dataset\cite{menze2014multimodal}, which provides comprehensive multi-modal MRI scans accompanied by expert-annotated ground truths. We strategically select T1-weighted contrast-enhanced (T1ce) and FLAIR as our representative source modalities. Clinically, T1ce strictly highlights the gadolinium-enhancing tumor core (ET) and necrotic regions (NCR), FLAIR exceptionally delineates the peritumoral edema (ED) spreading into healthy brain tissue by suppressing cerebrospinal fluid signals. Fusing these two modalities provides the most critical anatomical and pathological boundaries required for holistic brain tumor analysis.

\paragraph{\textbf{Evaluation Metrics.}} We rigorously quantify the segmentation performance using the Dice Similarity Coefficient and Intersection over Union (IoU) These metrics are computed across three distinct pathological sub-regions: NCR, ED, and ET, alongside their global averages to evaluate overall delineation accuracy. To ensure a strictly fair comparison, we evaluate our method against two distinct categories of state-of-the-art (SOTA) baselines:
(1) \textbf{Disjoint Fusion-Segmentation Pipelines:} We construct pipelines using recent SOTA image fusion models, including BSAFusion\cite{li2025bsafusion}, MATR\cite{tang2022matr}, DDFM\cite{zhao2023ddfm}, CDDFuse\cite{zhao2023cddfuse}, TFS-Diff\cite{xu2024simultaneous}, and FusionMamba\cite{xie2024fusionmamba}. To eliminate architecture-induced bias, the resulting fused images from all these methods are evaluated using the exact same segmentation backbone and training schedules as ours.
(2) \textbf{Direct Multi-Channel Methods:} We compare against heavily parameterized multi-channel segmentation models that directly consume all stacked raw modalities, including the nnU-Net\cite{isensee2021nnu}, TransUNet\cite{chen2021transunet}, CENet\cite{bozorgpour2025cenet} and VM-UNet\cite{ruan2024vm}. All models are strictly restricted to 2D slice inputs and 2 modalities.

\begin{figure}
    \centering
    \includegraphics[width=\linewidth]{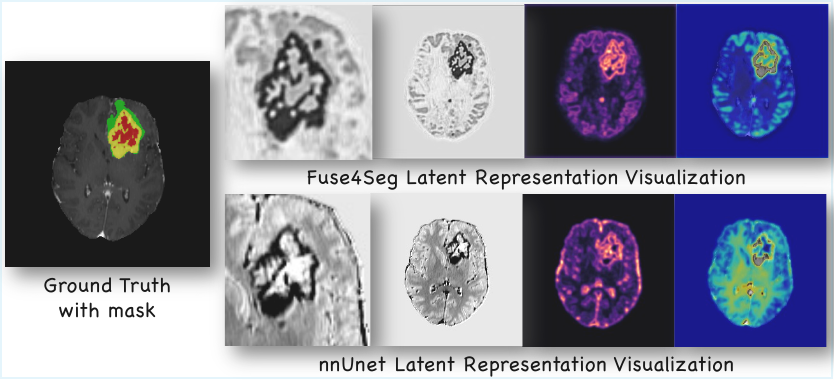}
    \caption{Visual interpretability comparison. Unlike the abstract, uninterpretable latent feature maps of traditional multi-channel networks (e.g., nnUNet), our Fuse4Seg explicitly bottleneck multi-modal information into a highly readable, task-driven fused prior ($x_f$).}
    \label{fig:interpretability}
\end{figure}

\paragraph{\textbf{Quantitative Comparison.}}
As shown in Table \ref{tab:seg_results}, Fuse4Seg significantly outperforms all disjoint fusion-segmentation pipelines, demonstrating that static visual priors are suboptimal for semantic tasks. More impressively, despite compressing multi-modal inputs into a single-channel prior, our task-driven framework achieves the highest overall Mean Dice, surpassing vanilla segmentation SOTAs. Excelling in delineating the critical Necrotic Core (NCR), Peritumoral Edema (ED) and Enhancing Tumor (ET), Fuse4Seg's highly optimized semantic prior establishes a new SOTA for efficient, image fusion-guided segmentation.

\begin{table*}[htbp]
\centering
\caption{Quantitative comparison of medical image fusion performance. The best results are highlighted in \textbf{bold}, and the second-best are \underline{underlined}. The symbol $\uparrow$ indicates that higher values denote better performance.}
\label{tab:fusion_results}
\resizebox{\textwidth}{!}{
\begin{tabular}{l|ccccccc|ccccccc}
\toprule
\multirow{2}{*}{\textbf{Method}} & \multicolumn{7}{c|}{\textbf{Harvard MRI-SPECT}} & \multicolumn{7}{c}{\textbf{Harvard MRI-PET}} \\
\cmidrule(lr){2-8} \cmidrule(lr){9-15}
& \textbf{EN $\uparrow$} & \textbf{SD $\uparrow$} & \textbf{MI $\uparrow$} & \textbf{PSNR $\uparrow$} & \textbf{SF $\uparrow$} & \textbf{$Q^{AB/F} \uparrow$} & \textbf{SSIM $\uparrow$} & \textbf{EN $\uparrow$} & \textbf{SD $\uparrow$} & \textbf{MI $\uparrow$} & \textbf{PSNR $\uparrow$} & \textbf{SF $\uparrow$} & \textbf{$Q^{AB/F} \uparrow$} & \textbf{SSIM $\uparrow$} \\
\midrule
MATR & 6.812 & 43.15 & 2.754 & 18.24 & 14.32 & 0.612 & 0.724 & 6.715 & 42.10 & 2.650 & 17.95 & 13.85 & 0.595 & 0.710 \\
BSAFusion & 6.905 & 44.52 & 2.915 & 19.55 & 15.01 & 0.638 & 0.741 & 6.820 & 43.65 & 2.810 & 19.10 & 14.65 & 0.615 & 0.725 \\
CDDFuse & 7.142 & 46.88 & 3.210 & 21.34 & 16.45 & 0.685 & 0.778 & 7.050 & 45.90 & 3.105 & 20.85 & 15.95 & 0.665 & 0.760 \\
DDFM & 7.351 & 49.21 & 3.842 & 20.12 & 18.22 & 0.671 & 0.752 & 7.250 & 48.15 & 3.750 & 19.85 & 17.85 & 0.655 & 0.740 \\
FusionMamba & 7.415 & 50.14 & 3.755 & \underline{22.05} & 17.54 & \underline{0.705} & \underline{0.792} & 7.310 & 49.25 & 3.650 & \underline{21.55} & 17.10 & \underline{0.685} & \underline{0.775} \\
TFS-Diff & \underline{7.450} & \textbf{51.36} & \textbf{4.015} & 21.88 & \underline{18.95} & 0.682 & 0.765 & \underline{7.380} & \textbf{50.60} & \underline{3.818} & 21.40 & \underline{18.50} & 0.660 & 0.750 \\
\midrule
\textbf{Ours (Fuse4Seg)} & \textbf{7.522} & \underline{50.85} & \underline{3.912} & \textbf{24.15} & \textbf{19.85} & \textbf{0.742} & \textbf{0.825} & \textbf{7.450} & \underline{49.85} & \textbf{3.920} & \textbf{23.85} & \textbf{19.25} & \textbf{0.728} & \textbf{0.812} \\
\bottomrule
\end{tabular}
}
\end{table*}

\begin{figure*}[t]
    \centering
    \includegraphics[width=\linewidth]{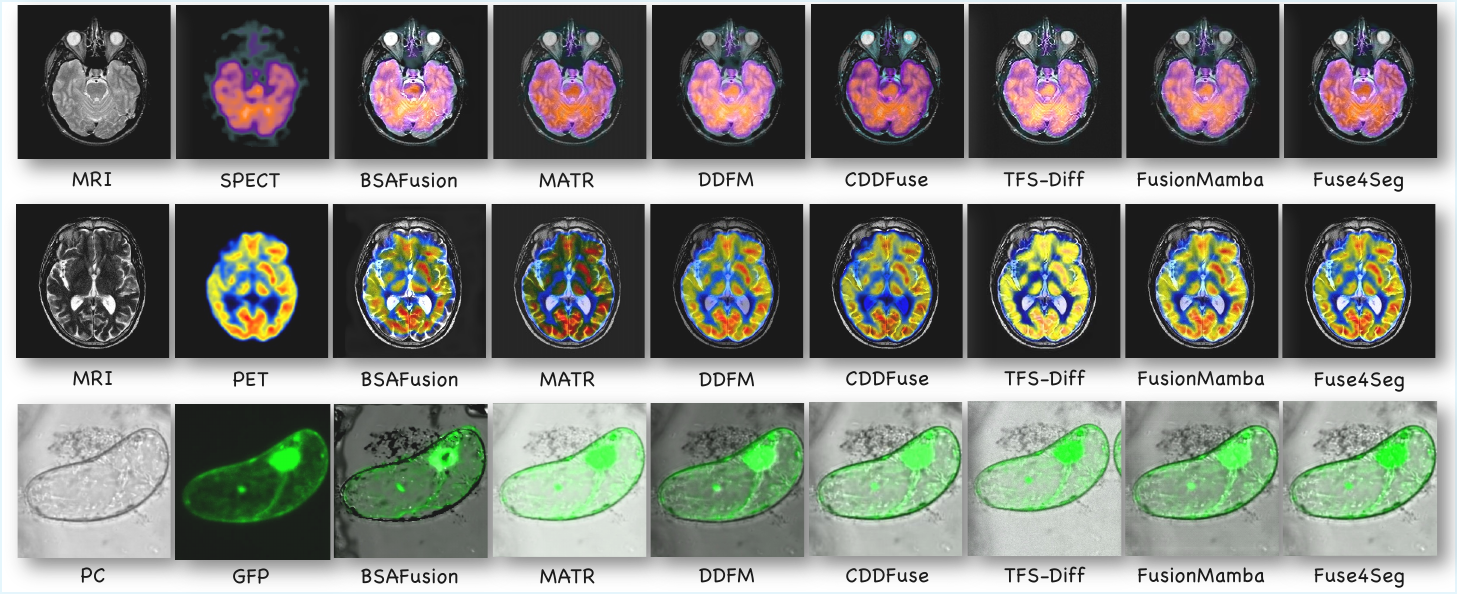} 
    \caption{Qualitative comparison with other SOTA methods on Harvard dataset MRI-SPECT, MRI-PET pairs, and GPF database GPF-PC pairs.}
    \label{fig:fusion}
\end{figure*}

\paragraph{\textbf{Qualitative Comparison.}} Visual inspections (Fig. \ref{fig:qualitative}) further validate the clinical superiority of our framework. As shown, disjoint pipelines (e.g., MATR+Seg) severely smear tumor boundaries, producing blocky artifacts and anatomical hallucinations. Furthermore, while direct multi-channel models (nnU-Net, VM-UNet) capture the general lesion area, they frequently over- or under-segment the highly irregular necrotic core (red) and enhancing tumor (yellow). In contrast, Fuse4Seg accurately delineates intricate micro-vascular boundaries and complex sub-region topologies with a precision that closely mirrors the expert-annotated Ground Truth.

\subsection{Visual Interpretability}
\label{sec:interpretability}

Vanilla multi-channel models are inherently ``black-box,'' yielding abstract latent features that lack intuitive physical meaning.

In contrast, Fuse4Seg offers a ``glass-box'' paradigm by bottlenecking multi-modal inputs into a single-channel physical prior ($x_f$). As shown in Fig. \ref{fig:interpretability}, unlike the uninterpretable activations of standard networks (e.g., nnU-Net), our fused prior explicitly bridges human and machine vision. Driven by downstream semantic gradients, the fusion leader dynamically suppresses redundant background tissue while enhancing contrast in pathological regions. Crucially, strictly regularized by $\mathcal{L}_{grad}$ and $\mathcal{L}_{recon}$, this enhancement avoids adversarial noise, preserving sharp, anatomically faithful boundaries. This explicit interpretability allows clinicians to effectively ``see what the network sees,'' verifying that predictions stem from valid biological structures rather than artifactual correlations, thereby fostering essential trust in the diagnostic pipeline.

\begin{table*}[htbp]
\centering
\caption{Ablation study on the core components of our Fuse4Seg framework. Performance is evaluated comprehensively across all tumor sub-regions (NCR, ED, ET) using Dice and IoU metrics.}
\label{tab:ablation}
\resizebox{\textwidth}{!}{
\begin{tabular}{lccc|cc|cc|cc|cc}
\toprule
\multirow{2}{*}{\textbf{Model Variants}} & \multicolumn{3}{c|}{\textbf{Components}} & \multicolumn{2}{c|}{\textbf{NCR}} & \multicolumn{2}{c|}{\textbf{ED}} & \multicolumn{2}{c|}{\textbf{ET}} & \multicolumn{2}{c}{\textbf{Average}} \\
\cmidrule(lr){2-4} \cmidrule(lr){5-6} \cmidrule(lr){7-8} \cmidrule(lr){9-10} \cmidrule(lr){11-12}
& \textbf{Bi-level} & $\mathcal{L}_{decomp}$ & $\mathcal{L}_{grad}$ & \textbf{Dice $\uparrow$} & \textbf{IoU $\uparrow$} & \textbf{Dice $\uparrow$} & \textbf{IoU $\uparrow$} & \textbf{Dice $\uparrow$} & \textbf{IoU $\uparrow$} & \textbf{Mean Dice $\uparrow$} & \textbf{Mean IoU $\uparrow$} \\
\midrule
Baseline (Disjoint) & $\times$ & $\times$ & $\times$  & 0.795 & 0.690 & 0.815 & 0.715 & 0.850 & 0.749 & 0.820 & 0.718 \\
Variant A & \checkmark & $\times$ & $\times$ & 0.840 & 0.745 & 0.855 & 0.755 & 0.900 & 0.780 & 0.865 & 0.760 \\
Variant B & \checkmark & \checkmark & $\times$ & 0.865 & 0.775 & 0.880 & 0.785 & 0.919 & 0.825 & 0.888 & 0.795 \\
\textbf{Ours (Full)} & \checkmark & \checkmark & \checkmark & \textbf{0.884} & \textbf{0.807} & \textbf{0.908} & \textbf{0.833} & \textbf{0.937} & \textbf{0.884} & \textbf{0.910} & \textbf{0.841} \\
\bottomrule
\end{tabular}
}
\end{table*}

\subsection{Medical Image Fusion Results}
\paragraph{\textbf{Datasets.}} 
To rigorously validate the generalization capability of our fusion module beyond the BraTS segmentation task, we extend our evaluation to two fusion benchmarks: the Harvard dataset\cite{johnson_becker_harvard_medical} and the GFP database\cite{jic_gfp_dataset}. From the Harvard dataset, we use MRI-SPECT and MRI-PET modality pairs to assess the framework's ability to seamlessly integrate high-resolution macroscopic anatomical structures (MRI) with functional metabolic signals (PET/SPECT). Furthermore, to evaluate robustness across drastically different physical scales, we employ Green Fluorescent Protein and Phase Contrast (GFP-PC) image pairs from the GFP database. Training and evaluating of our fusion module on these diverse modalities demonstrates that our frequency-decoupled architecture is serves as a highly adaptable, generalized prior generator for both macroscopic and microscopic medical image fusion.

\paragraph{\textbf{Evaluation Metrics.}}
We benchmark our \textit{Fuse4Seg} against seven state-of-the-art (SOTA) medical image fusion methods, spanning CNN-based, Transformer-based, Diffusion-based, and Mamba-based architectures. The comparative baselines include MATR\cite{tang2022matr}, BSAFusion\cite{li2025bsafusion}, CDDFuse\cite{zhao2023cddfuse}, DDFM\cite{zhao2023ddfm}, TFS-Diff\cite{xu2024simultaneous} and FusionMamba\cite{xie2024fusionmamba}. To comprehensively evaluate the fusion quality, we select seven widely-adopted objective metrics: Entropy (EN)\cite{roberts2008assessment}, Standard Deviation (SD)\cite{eskicioglu2002image}, Mutual Information (MI)\cite{ma2019infrared}, Peak Signal-to-Noise Ratio (PSNR)\cite{hore2010image}, Spatial Frequency (SF)\cite{ma2019infrared}, Edge Preservation ($Q^{AB/F}$)\cite{ma2019infrared}, and Structural Similarity (SSIM)\cite{wang2004image}.

\paragraph{\textbf{Quantitative Comparison.}} As Table \ref{tab:fusion_results} reports, explicitly guided by $\mathcal{L}_{grad}$, Fuse4Seg comprehensively dominates structural and fidelity metrics (PSNR, SF, $Q^{AB/F}$, SSIM), proving it preserves sharp anatomical edges without structural distortion. While generative pipelines like TFS-Diff marginally lead in global statistics (EN, SD), this is anatomically expected; diffusion models often inject synthetic noise that artificially inflates entropy. In contrast, Fuse4Seg tightly bounds the feature space to retain authentic clinical readability, yielding highly robust fused priors across MRI-SPECT and MRI-PET.

\paragraph{\textbf{Qualitative Comparison.}} Visual inspections (Fig. \ref{fig:fusion}) corroborate our quantitative superiority across multi-scale modalities. As observed, baseline methods frequently struggle to balance structural and functional information. For instance, MATR and BSAFusion introduce severe color distortions and contrast degradation, while approaches like TFS-Diff tend to over-saturate functional signals, inadvertently masking underlying anatomical boundaries. In contrast, tightly constrained by our spatial gradient loss and physical reconstruction anchor, Fuse4Seg seamlessly integrates the high-frequency structural details of MRI and PC images with the vibrant functional representations of PET, SPECT, and GFP modalities. Consequently, our method produces sharp, artifact-free fused images with maximal clinical interpretability.

\subsection{Ablation Study}
To validate the efficacy of our proposed components, we conduct a comprehensive ablation study, incrementally adding the Cooperative Bi-level Optimization (Bi-level), Frequency Decomposition Loss ($\mathcal{L}_{decomp}$), and Spatial Gradient Loss ($\mathcal{L}_{grad}$) to a disjoint AutoEncoder baseline.

\paragraph{\textbf{Effect of Bi-level Optimization.}} 
Introducing the alternating bi-level loop significantly boosts overall segmentation performance, particularly for the Enhancing Tumor (ET). This confirms that allowing semantic gradients from the downstream follower to directly backpropagate successfully forces the fusion leader to highlight critical semantic regions, generating a task-aware prior rather than a naive physical blend.

\paragraph{\textbf{Effect of Frequency Decomposition ($\mathcal{L}_{decomp}$).}} 
To counter potential feature entanglement, we incorporate $\mathcal{L}_{decomp}$. Enforcing a strict separation between modality-shared anatomical structures and modality-specific high-frequency textures filters out redundant physical noise. This provides a significantly cleaner semantic prior, which is particularly vital for elevating the segmentation accuracy of the intricate Necrotic Core (NCR).

\paragraph{\textbf{Effect of Spatial Gradient Loss ($\mathcal{L}_{grad}$).}} 
While previous variants capture broad semantics, they lack explicit physical boundary constraints, occasionally over-smoothing local textures. Adding $\mathcal{L}_{grad}$ explicitly forces the fused image to inherit sharp diagnostic boundaries from the source modalities. This physical regularity synergistically maximizes our segmentation performance, proving crucial for delineating the diffuse, highly irregular boundaries of the Peritumoral Edema (ED) and culminating in our state-of-the-art results.

\section{Conclusion}
In this paper, we propose \textit{Fuse4Seg}, reformulating multi-modal fusion as a cooperative bi-level optimization problem to bridge the gap with high-level semantic segmentation. By abandoning disjoint, human-perception-driven pipelines, our method dynamically compresses diverse modalities into a highly optimized, task-aware single-channel prior. Extensive experiments demonstrate that Fuse4Seg achieves state-of-the-art fusion fidelity while remarkably surpassing contemporary dual-channel segmentation SOTAs. Ultimately, by bottlenecking abstract features into a readable ``glass-box'' physical image, our framework provides crucial visual interpretability, establishing a robust, transparent baseline for task-driven medical diagnosis.

\section*{Acknowledgments}
This work was supported in part by the Guangdong Provincial Key Laboratory of IRADS (2022B1212010006), in part by the Guangdong Higher Education Upgrading Plan (2021–2025), in part by the Guangdong and Hong Kong Universities “1+1+1” Joint Research Collaboration Scheme, in part by the National Key Research and Development Program of China (2022ZD0117700), in part by the National Natural Science Foundation of China (62325204), and in part by the MSAI program of Northwestern University. The authors would like to thank Zhenghao Wu's insightful discussions and feedback on the manuscript.

{\small
\bibliographystyle{ieee_fullname}
\bibliography{egbib}

@String(AAAI  = {AAAI})

@String(ICASSP=	{ICASSP})

@String(ICME  = {Int. Conf. Multimedia and Expo})

@String(ICME  =	{ICME})

@inproceedings{li2025bsafusion,
  title={Bsafusion: A bidirectional stepwise feature alignment network for unaligned medical image fusion},
  author={Li, Huafeng and Su, Dayong and Cai, Qing and Zhang, Yafei},
  booktitle={Proceedings of the AAAI conference on artificial intelligence},
  volume={39},
  number={5},
  pages={4725--4733},
  year={2025}
}

@article{tang2022matr,
  title={MATR: Multimodal medical image fusion via multiscale adaptive transformer},
  author={Tang, Wei and He, Fazhi and Liu, Yu and Duan, Yansong},
  journal={IEEE Transactions on Image Processing},
  volume={31},
  pages={5134--5149},
  year={2022},
  publisher={IEEE}
}

@inproceedings{zhao2023ddfm,
  title={DDFM: Denoising diffusion model for multi-modality image fusion},
  author={Zhao, Zixiang and Bai, Haowen and Zhu, Yuanzhi and Zhang, Jiangshe and Xu, Shuang and Zhang, Yulun and Zhang, Kai and Meng, Deyu and Timofte, Radu and Van Gool, Luc},
  booktitle={Proceedings of the IEEE/CVF international conference on computer vision},
  pages={8082--8093},
  year={2023}
}

@inproceedings{zhao2023cddfuse,
  title={Cddfuse: Correlation-driven dual-branch feature decomposition for multi-modality image fusion},
  author={Zhao, Zixiang and Bai, Haowen and Zhang, Jiangshe and Zhang, Yulun and Xu, Shuang and Lin, Zudi and Timofte, Radu and Van Gool, Luc},
  booktitle={Proceedings of the IEEE/CVF conference on computer vision and pattern recognition},
  pages={5906--5916},
  year={2023}
}

@inproceedings{xu2024simultaneous,
  title={Simultaneous tri-modal medical image fusion and super-resolution using conditional diffusion model},
  author={Xu, Yushen and Li, Xiaosong and Jie, Yuchan and Tan, Haishu},
  booktitle={International conference on medical image computing and computer-assisted intervention},
  pages={635--645},
  year={2024},
  organization={Springer}
}

@article{xie2024fusionmamba,
  title={Fusionmamba: Dynamic feature enhancement for multimodal image fusion with mamba},
  author={Xie, Xinyu and Cui, Yawen and Tan, Tao and Zheng, Xubin and Yu, Zitong},
  journal={Visual Intelligence},
  volume={2},
  number={1},
  pages={37},
  year={2024},
  publisher={Springer}
}

@article{isensee2021nnu,
  title={nnU-Net: a self-configuring method for deep learning-based biomedical image segmentation},
  author={Isensee, Fabian and Jaeger, Paul F and Kohl, Simon AA and Petersen, Jens and Maier-Hein, Klaus H},
  journal={Nature methods},
  volume={18},
  number={2},
  pages={203--211},
  year={2021},
  publisher={Nature Publishing Group US New York}
}

@article{chen2021transunet,
  title={Transunet: Transformers make strong encoders for medical image segmentation},
  author={Chen, Jieneng and Lu, Yongyi and Yu, Qihang and Luo, Xiangde and Adeli, Ehsan and Wang, Yan and Lu, Le and Yuille, Alan L and Zhou, Yuyin},
  journal={arXiv preprint arXiv:2102.04306},
  year={2021}
}

@article{ruan2024vm,
  title={Vm-unet: Vision mamba unet for medical image segmentation},
  author={Ruan, Jiacheng and Li, Jincheng and Xiang, Suncheng},
  journal={ACM Transactions on Multimedia Computing, Communications and Applications},
  year={2024},
  publisher={ACM New York, NY}
}

@inproceedings{bozorgpour2025cenet,
  title={Cenet: Context enhancement network for medical image segmentation},
  author={Bozorgpour, Afshin and Kolahi, Sina Ghorbani and Azad, Reza and Hacihaliloglu, Ilker and Merhof, Dorit},
  booktitle={International Conference on Medical Image Computing and Computer-Assisted Intervention},
  pages={120--129},
  year={2025},
  organization={Springer}
}

@article{roberts2008assessment,
  title={Assessment of image fusion procedures using entropy, image quality, and multispectral classification},
  author={Roberts, J Wesley and Van Aardt, Jan A and Ahmed, Fethi Babikker},
  journal={Journal of Applied Remote Sensing},
  volume={2},
  number={1},
  pages={023522},
  year={2008},
  publisher={SPIE}
}

@article{eskicioglu2002image,
  title={Image quality measures and their performance},
  author={Eskicioglu, Ahmet M and Fisher, Paul S},
  journal={IEEE Transactions on communications},
  volume={43},
  number={12},
  pages={2959--2965},
  year={2002},
  publisher={IEEE}
}

@article{ma2019infrared,
  title={Infrared and visible image fusion methods and applications: A survey},
  author={Ma, Jiayi and Ma, Yong and Li, Chang},
  journal={Information fusion},
  volume={45},
  pages={153--178},
  year={2019},
  publisher={Elsevier}
}

@inproceedings{hore2010image,
  title={Image quality metrics: PSNR vs. SSIM},
  author={Hore, Alain and Ziou, Djemel},
  booktitle={2010 20th international conference on pattern recognition},
  pages={2366--2369},
  year={2010},
  organization={IEEE}
}

@article{wang2004image,
  title={Image quality assessment: from error visibility to structural similarity},
  author={Wang, Zhou and Bovik, Alan C and Sheikh, Hamid R and Simoncelli, Eero P},
  journal={IEEE transactions on image processing},
  volume={13},
  number={4},
  pages={600--612},
  year={2004},
  publisher={IEEE}
}

@article{james2014medical,
  title={Medical image fusion: A survey of the state of the art},
  author={James, Alex Pappachen and Dasarathy, Belur V},
  journal={Information fusion},
  volume={19},
  pages={4--19},
  year={2014},
  publisher={Elsevier}
}

@article{azam2022review,
  title={A review on multimodal medical image fusion: Compendious analysis of medical modalities, multimodal databases, fusion techniques and quality metrics},
  author={Azam, Muhammad Adeel and Khan, Khan Bahadar and Salahuddin, Sana and Rehman, Eid and Khan, Sajid Ali and Khan, Muhammad Attique and Kadry, Seifedine and Gandomi, Amir H},
  journal={Computers in biology and medicine},
  volume={144},
  pages={105253},
  year={2022},
  publisher={Elsevier}
}

@article{tan2025multimodal,
  title={Multimodal medical image fusion algorithm in the era of big data},
  author={Tan, Wei and Tiwari, Prayag and Pandey, Hari Mohan and Moreira, Catarina and Jaiswal, Amit Kumar},
  journal={Neural computing and applications},
  volume={37},
  number={28},
  pages={22995--23015},
  year={2025},
  publisher={Springer}
}

@inproceedings{guo2025dae,
  title={Dae-fuse: An adaptive discriminative autoencoder for multi-modality image fusion},
  author={Guo, Yuchen and Xu, Ruoxiang and Li, Rongcheng and Su, Weifeng},
  booktitle={2025 IEEE International Conference on Multimedia and Expo (ICME)},
  pages={1--6},
  year={2025},
  organization={IEEE}
}

@article{tang2025mask,
  title={Mask-DiFuser: A masked diffusion model for unified unsupervised image fusion},
  author={Tang, Linfeng and Li, Chunyu and Ma, Jiayi},
  journal={IEEE Transactions on Pattern Analysis and Machine Intelligence},
  year={2025},
  publisher={IEEE}
}

@inproceedings{liu2023multi,
  title={Multi-interactive feature learning and a full-time multi-modality benchmark for image fusion and segmentation},
  author={Liu, Jinyuan and Liu, Zhu and Wu, Guanyao and Ma, Long and Liu, Risheng and Zhong, Wei and Luo, Zhongxuan and Fan, Xin},
  booktitle={Proceedings of the IEEE/CVF international conference on computer vision},
  pages={8115--8124},
  year={2023}
}

@inproceedings{liu2022target,
  title={Target-aware dual adversarial learning and a multi-scenario multi-modality benchmark to fuse infrared and visible for object detection},
  author={Liu, Jinyuan and Fan, Xin and Huang, Zhanbo and Wu, Guanyao and Liu, Risheng and Zhong, Wei and Luo, Zhongxuan},
  booktitle={Proceedings of the IEEE/CVF conference on computer vision and pattern recognition},
  pages={5802--5811},
  year={2022}
}

@article{ma2020ddcgan,
  title={DDcGAN: A dual-discriminator conditional generative adversarial network for multi-resolution image fusion},
  author={Ma, Jiayi and Xu, Han and Jiang, Junjun and Mei, Xiaoguang and Zhang, Xiao-Ping},
  journal={IEEE Transactions on Image Processing},
  volume={29},
  pages={4980--4995},
  year={2020},
  publisher={IEEE}
}

@inproceedings{cao2022swin,
  title={Swin-unet: Unet-like pure transformer for medical image segmentation},
  author={Cao, Hu and Wang, Yueyue and Chen, Joy and Jiang, Dongsheng and Zhang, Xiaopeng and Tian, Qi and Wang, Manning},
  booktitle={European conference on computer vision},
  pages={205--218},
  year={2022},
  organization={Springer}
}

@inproceedings{huang2020unet,
  title={Unet 3+: A full-scale connected unet for medical image segmentation},
  author={Huang, Huimin and Lin, Lanfen and Tong, Ruofeng and Hu, Hongjie and Zhang, Qiaowei and Iwamoto, Yutaro and Han, Xianhua and Chen, Yen-Wei and Wu, Jian},
  booktitle={ICASSP 2020-2020 IEEE international conference on acoustics, speech and signal processing (ICASSP)},
  pages={1055--1059},
  year={2020},
  organization={Ieee}
}

@article{wang2024mamba,
  title={Mamba-unet: Unet-like pure visual mamba for medical image segmentation},
  author={Wang, Ziyang and Zheng, Jian-Qing and Zhang, Yichi and Cui, Ge and Li, Lei},
  journal={arXiv preprint arXiv:2402.05079},
  year={2024}
}

@article{chen2017deeplab,
  title={Deeplab: Semantic image segmentation with deep convolutional nets, atrous convolution, and fully connected crfs},
  author={Chen, Liang-Chieh and Papandreou, George and Kokkinos, Iasonas and Murphy, Kevin and Yuille, Alan L},
  journal={IEEE transactions on pattern analysis and machine intelligence},
  volume={40},
  number={4},
  pages={834--848},
  year={2017},
  publisher={IEEE}
}

@inproceedings{woo2023convnext,
  title={Convnext v2: Co-designing and scaling convnets with masked autoencoders},
  author={Woo, Sanghyun and Debnath, Shoubhik and Hu, Ronghang and Chen, Xinlei and Liu, Zhuang and Kweon, In So and Xie, Saining},
  booktitle={Proceedings of the IEEE/CVF conference on computer vision and pattern recognition},
  pages={16133--16142},
  year={2023}
}

@article{vaswani2017attention,
  title={Attention is all you need},
  author={Vaswani, Ashish and Shazeer, Noam and Parmar, Niki and Uszkoreit, Jakob and Jones, Llion and Gomez, Aidan N and Kaiser, {\L}ukasz and Polosukhin, Illia},
  journal={Advances in neural information processing systems},
  volume={30},
  year={2017}
}

@article{ardizzone2018analyzing,
  title={Analyzing inverse problems with invertible neural networks},
  author={Ardizzone, Lynton and Kruse, Jakob and Wirkert, Sebastian and Rahner, Daniel and Pellegrini, Eric W and Klessen, Ralf S and Maier-Hein, Lena and Rother, Carsten and K{\"o}the, Ullrich},
  journal={arXiv preprint arXiv:1808.04730},
  year={2018}
}

@article{azad2024medical,
  title={Medical image segmentation review: The success of u-net},
  author={Azad, Reza and Aghdam, Ehsan Khodapanah and Rauland, Amelie and Jia, Yiwei and Avval, Atlas Haddadi and Bozorgpour, Afshin and Karimijafarbigloo, Sanaz and Cohen, Joseph Paul and Adeli, Ehsan and Merhof, Dorit},
  journal={IEEE Transactions on Pattern Analysis and Machine Intelligence},
  volume={46},
  number={12},
  pages={10076--10095},
  year={2024},
  publisher={IEEE}
}

@article{muller2022towards,
  title={Towards a guideline for evaluation metrics in medical image segmentation},
  author={M{\"u}ller, Dominik and Soto-Rey, I{\~n}aki and Kramer, Frank},
  journal={BMC research notes},
  volume={15},
  number={1},
  pages={210},
  year={2022},
  publisher={Springer}
}

@article{wang2022medical,
  title={Medical image segmentation using deep learning: A survey},
  author={Wang, Risheng and Lei, Tao and Cui, Ruixia and Zhang, Bingtao and Meng, Hongying and Nandi, Asoke K},
  journal={IET image processing},
  volume={16},
  number={5},
  pages={1243--1267},
  year={2022},
  publisher={Wiley Online Library}
}

@article{zhao2020didfuse,
  title={DIDFuse: Deep image decomposition for infrared and visible image fusion},
  author={Zhao, Zixiang and Xu, Shuang and Zhang, Chunxia and Liu, Junmin and Li, Pengfei and Zhang, Jiangshe},
  journal={arXiv preprint arXiv:2003.09210},
  year={2020}
}

@article{yi2024diff,
  title={Diff-IF: Multi-modality image fusion via diffusion model with fusion knowledge prior},
  author={Yi, Xunpeng and Tang, Linfeng and Zhang, Hao and Xu, Han and Ma, Jiayi},
  journal={Information Fusion},
  volume={110},
  pages={102450},
  year={2024},
  publisher={Elsevier}
}

@article{menze2014multimodal,
  title={The multimodal brain tumor image segmentation benchmark (BRATS)},
  author={Menze, Bjoern H and Jakab, Andras and Bauer, Stefan and Kalpathy-Cramer, Jayashree and Farahani, Keyvan and Kirby, Justin and Burren, Yuliya and Porz, Nicole and Slotboom, Johannes and Wiest, Roland and others},
  journal={IEEE transactions on medical imaging},
  volume={34},
  number={10},
  pages={1993--2024},
  year={2014},
  publisher={IEEE}
}

@misc{johnson_becker_harvard_medical,
  author       = {Johnson, B. A. and Becker, J. A.},
  howpublished = {\url{http://www.med.harvard.edu/AANLIB/home.html}},
}

@misc{jic_gfp_dataset,
  title        = {GFP Image Dataset},
  howpublished = {\url{https://data.jic.ac.uk/Gfp/}},
}

@inproceedings{zhao2024equivariant,
  title={Equivariant multi-modality image fusion},
  author={Zhao, Zixiang and Bai, Haowen and Zhang, Jiangshe and Zhang, Yulun and Zhang, Kai and Xu, Shuang and Chen, Dongdong and Timofte, Radu and Van Gool, Luc},
  booktitle={Proceedings of the IEEE/CVF conference on computer vision and pattern recognition},
  pages={25912--25921},
  year={2024}
}

@article{liu2021learning,
  title={Learning a deep multi-scale feature ensemble and an edge-attention guidance for image fusion},
  author={Liu, Jinyuan and Fan, Xin and Jiang, Ji and Liu, Risheng and Luo, Zhongxuan},
  journal={IEEE Transactions on Circuits and Systems for Video Technology},
  volume={32},
  number={1},
  pages={105--119},
  year={2021},
  publisher={IEEE}
}
}

\end{document}